\documentclass{article}
\usepackage{ijcai16}
\usepackage{graphicx}
\usepackage{multirow}
\usepackage{times}
\usepackage{amssymb,amsfonts}
\title{Attention-Based Convolutional Neural Network\\ for Machine Comprehension}

\author{Wenpeng Yin, Sebastian Ebert, Hinrich Sch\"{u}tze \\ 
University of Munich, Germany \\
wenpeng, ebert@cis.lmu.de}

\def\passage{document}
\def\snippet{snippet}

\newcounter{notecounter}

\newcommand{\enoteson}{\long\gdef\enote##1##2{{

\stepcounter{notecounter}
\large\bf
\hspace{1cm}\arabic{notecounter} $<<<$ ##1: ##2 $>>>$\hspace{1cm}}}}
\enoteson
\def\seclabel#1{\label{sec:#1}\label{p:#1}}
\def\eqref#1{Eq.~\ref{eqn:#1}}

\begin{document}

\maketitle

\begin{abstract}
Understanding open-domain text is one of the primary
challenges in natural language processing (NLP). Machine
comprehension benchmarks evaluate the system's ability to understand
text based on
the text content only. In this work, we investigate machine
comprehension on MCTest, a question answering (QA) benchmark. Prior work
is mainly based on feature engineering approaches. We come
up with a neural network framework, named hierarchical
attention-based convolutional neural network (HABCNN), to
address this task without any manually designed
features. Specifically, we explore HABCNN for this task by
two routes, one is through traditional joint modeling
of \passage{}, question and answer, one is through textual
entailment. HABCNN employs an attention mechanism to detect key
phrases, key sentences and key \snippet{}s that are relevant to
answering the question. Experiments show that
HABCNN outperforms prior deep learning approaches
by a big margin.
\end{abstract}

\section{Introduction}
Endowing machines with the ability to understand natural language
is a long-standing goal in NLP and holds the promise of
revolutionizing the way in which people interact with
machines and retrieve information. Richardson et al. \shortcite{richardson2013mctest}
proposed the task of machine comprehension, along with
MCTest, a question answering 
dataset for evaluation. The ability of the machine to
understand text is evaluated 
by posing a
series of questions, where the answer to each question can
be found only in the associated text. Solutions typically
focus on some semantic interpretation of the text, possibly
with some form of probabilistic or logic inference, 
to answer the question. Despite intensive recent work
\cite{weston2014memory,weston2015towards,hermann2015teaching,sachan2015learning},
the problem is far from solved.

\begin{figure}[t]
\setlength{\abovecaptionskip}{-3pt} 
\setlength{\belowcaptionskip}{-10pt} 
\centering
\includegraphics[width=8.5cm]{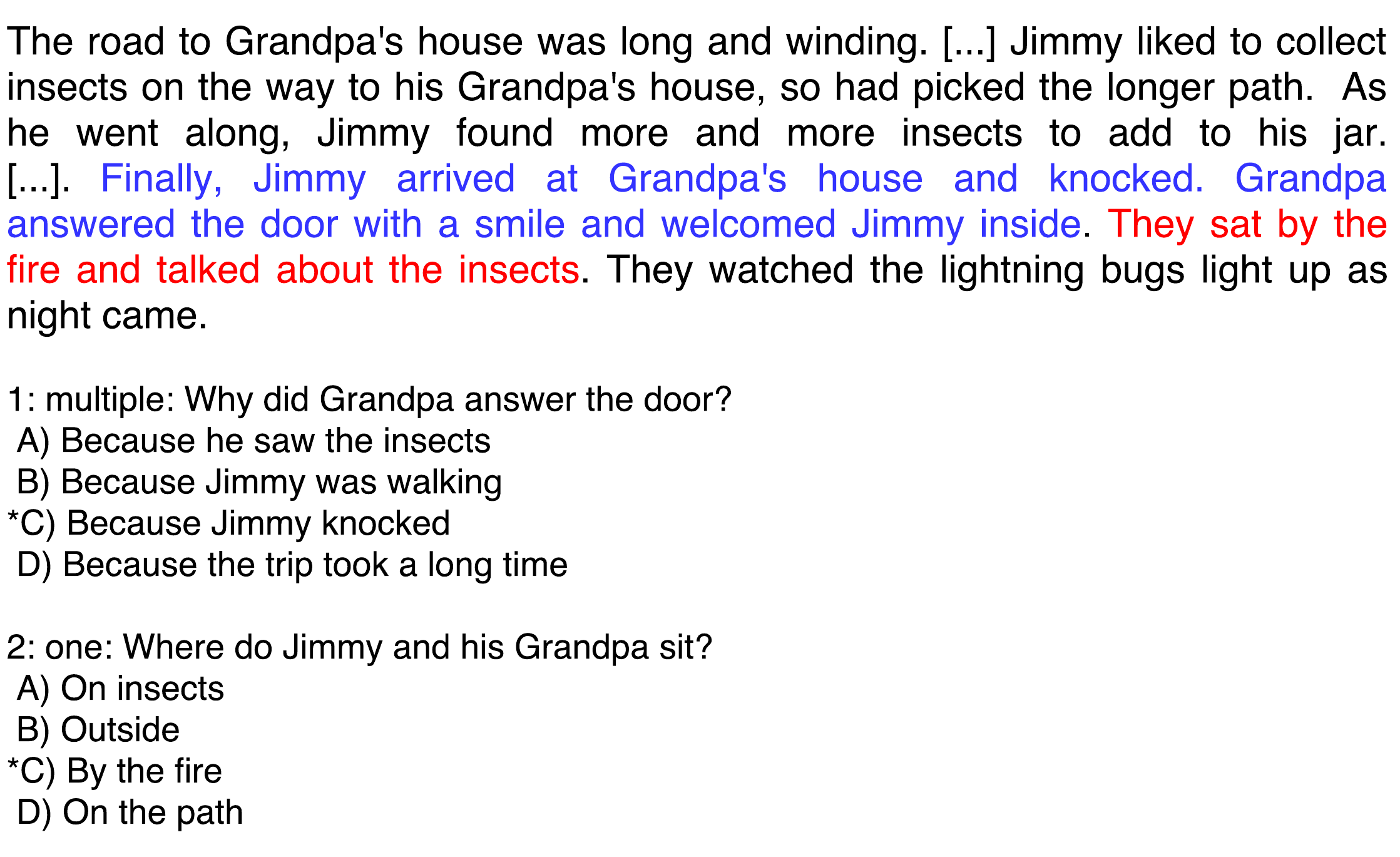}
\caption{One example with 2 out of 4 questions in the MCTest. ``*" marks correct answer.} \label{fig:example}
\end{figure}

Machine comprehension is an open-domain question-answering
problem which contains factoid questions, but the answers
can be derived by extraction or induction of key
clues. Figure \ref{fig:example} shows one example in
MCTest. Each example consists of one \passage{}, four
associated questions; each question is followed by four
answer candidates in which only one is correct. Questions in
MCTest have two categories: ``one'' and ``multiple''. The
label means \emph{one} or \emph{multiple} sentences from
the \passage{} are required to answer this question. To
correctly answer the first question in the example, the two
blue sentences are required; for the second question
instead, only the red sentence can help.  The following
observations hold for the whole MCTest. (i) Most of the
sentences in the \passage{} are irrelavent for a given
question. It hints that we need to pay attention to just
some key regions. (ii) The answer candidates can be flexible
text in length and abstraction level, and probably do not
appear in the \passage{}. For example, candidate B for the
second question is ``outside'', which is one word and does
not exist in the \passage{}, while the answer candidates for
the first question are longer texts with some
auxiliary words like ``Because'' in the text. This requires
our system to handle flexible texts via \emph{extraction} as
well as \emph{abstraction}. (iii) Some questions require
multiple sentences to infer the answer, and those vital
sentences mostly appear close to each other (we call
them \emph{\snippet{}}). Hence, our system should be able to
make a choice or compromise between potential
single-sentence clue and \snippet{} clue.

Prior work on this task  is mostly based on feature engineering. This work, instead, takes the lead in presenting a deep neural network based approach without any linguistic features involved.  

Concretely, we propose HABCNN, a hierarchical attention-based
convolutional neural network, to
address this task in two roadmaps. In the first one, 
we
project the document 
in two different ways, one based on question-attention,
one based on answer-attention and then compare the two projected document representations to
determine whether the answer matches the question.
In the second one, every question-answer pair is reformatted into a statement,
then the whole task is treated through textual entailment.

In both roadmaps, convolutional neural network (CNN) is
explored to model all types of text. As human beings usually
do for such a QA task, our model is expected to be able to
detect the key \snippet{}s, key sentences, and key words or
phrases in the \passage{}. In order to detect those
informative parts required by questions, we explore an
attention mechanism to model the \passage{} so that its
representation contains required information intensively. In
practice, instead of imitating human beings in QA task
top-down, our system models the \passage{} bottom-up,
through accumulating the most relevant information from word
level to \snippet{} level.

Our approach is novel in three aspects.
(i) A \passage{} is
modeled by a hierarchical CNN for different granularity,
from word to sentence level, then from sentence
to \snippet{} level. The reason of choosing a CNN rather
than other sequence models like recurrent neural
network \cite{mikolov2010recurrent}, long short-term memory
unit (LSTM \cite{hochreiter1997long}), gated recurrent unit
(GRU \cite{cho2014learning}) etc, is that we argue CNNs are
more suitable to detect the key sentences within \passage{}s
and key phrases within sentences. Considering again the
second question in Figure \ref{fig:example}, the original
sentence ``They sat by the fire and talked about he
insects'' has more information than required, i.e, we do not
need to know ``they talked about the insects''. Sequence
modeling neural networks usually model the sentence meaning by
accumulating the whole sequence. CNNs, with
convolution-pooling steps, are supposed to detect some
prominent features no matter where the features come from.
(ii) In the example in Figure \ref{fig:example}, apparently
not all sentences are required given a question, and usually
different \snippet{}s are required by different
questions. Hence, the same \passage{} should have different
representations based on what the question is. To this end,
attentions are incorporated into the hierarchical
CNN to guide the learning of dynamic \passage{}
representations which closely match the information
requirements by questions. (iii) Document representations
at sentence and snippet levels both are informative for
the question, a highway network is developed to combine
them, enabling our system to make a flexible tradeoff.

Overall, we make three contributions. (i) We
present a hierarchical attention-based CNN system
``HABCNN''. It is, to our knowledge, the first deep learning
based system for this MCTest task. (ii) Prior \passage{} modeling
systems based on deep neural networks mostly generate
generic representation, this work is the first to
incorporate attention so that \passage{} representation is
biased towards the question requirement. (iii) Our HABCNN
systems outperform other deep learning competitors by big
margins.

\begin{figure}[t]
\setlength{\abovecaptionskip}{-3pt} 
\setlength{\belowcaptionskip}{-10pt}
\centering
\includegraphics[width=5.5cm]{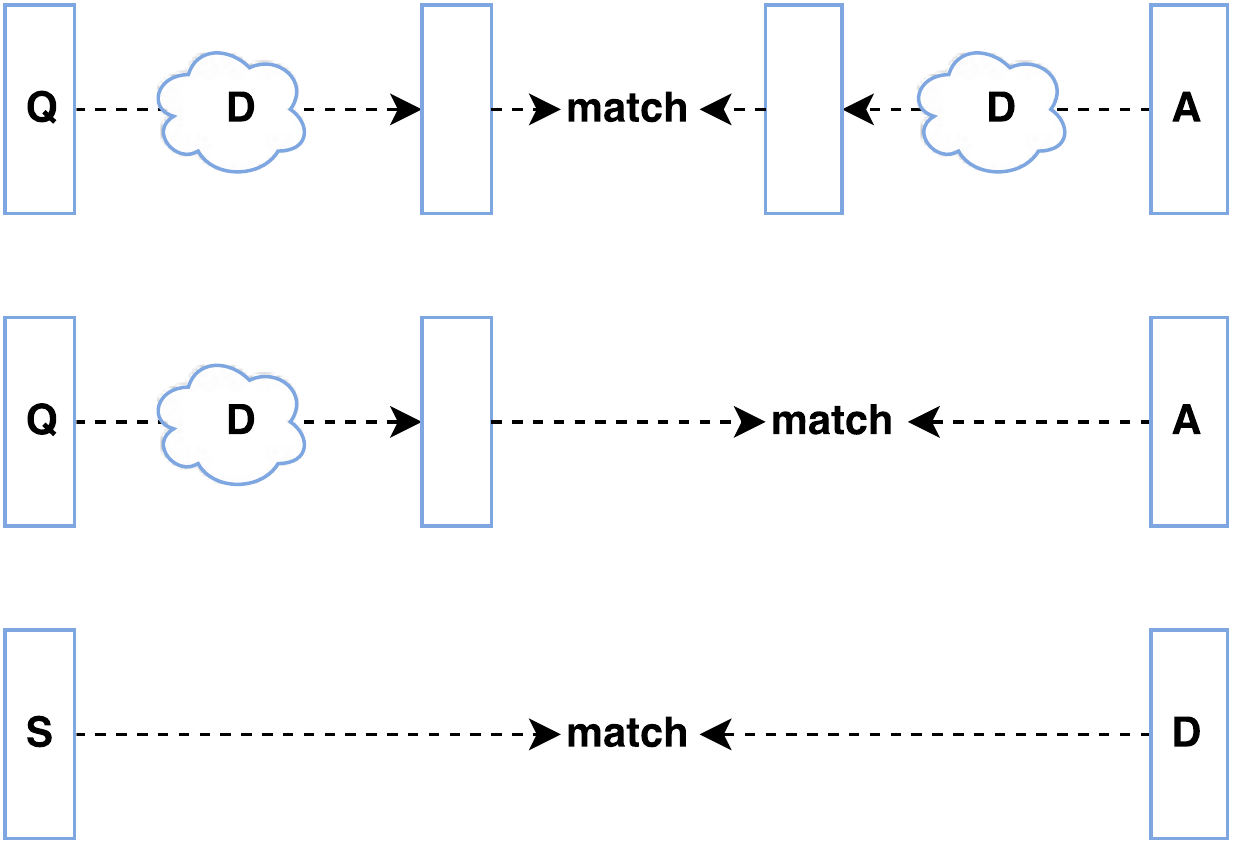}
\caption{Illustrations of HABCNN-QAP (top), HABCHH-QP (middle) and HABCNN-TE (bottom). Q, A, S: question, answer, statement; D: \passage{}} \label{fig:routes}
\end{figure}
\begin{figure*}[t]
\setlength{\abovecaptionskip}{-3pt} 
\setlength{\belowcaptionskip}{-10pt}
\centering
\includegraphics[width=14cm]{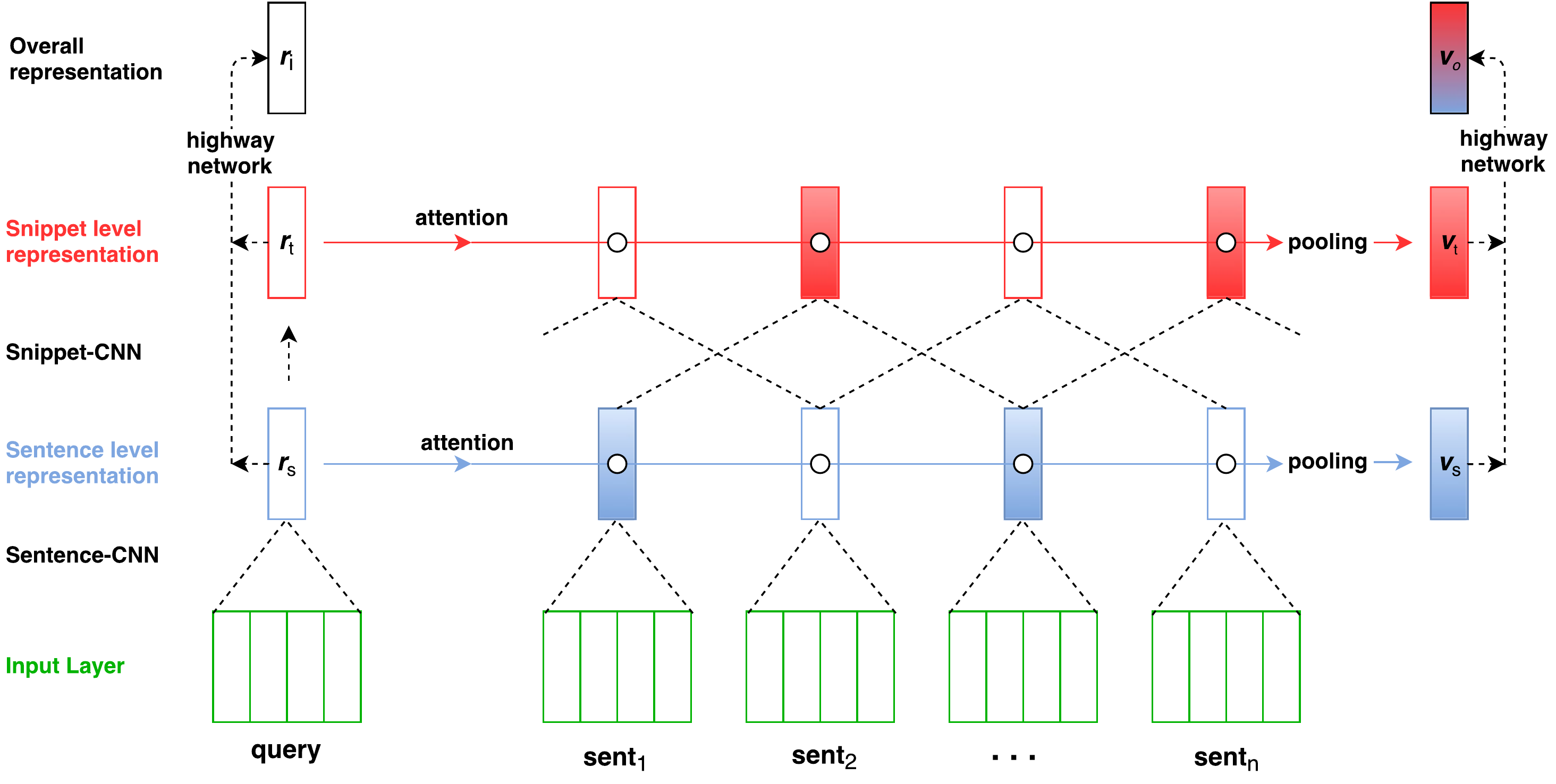}
\caption{HABCNN. Feature maps for phrase representations $\mathbf{p}_i$ and the max pooling steps that create sentence representations out of phrase representations are omitted for simplification. Each snippet covers three sentences in snippet-CNN. Symbols $\circ$ mean cosine similarity calculation.} \label{fig:abhcnn}
\end{figure*}

\section{Related Work}\label{sec:relatedwork}
Existing systems for MCTest task are mostly based on
manually engineered features. Representative work
includes \cite{narasimhan2015machine,sachan2015learning,wang2015machine,smith2015EMNLP}. In
these works, a common route is first to define a regularized
loss function based on assumed feature vectors, then the
effort focuses on designing effective features based on
various rules. Even though these researches are
groundbreaking for this task, their 
flexibility and their capacity for generalization are limited.

Deep learning based approaches appeal to increasing interest
in analogous tasks. Weston et
al., \shortcite{weston2014memory} introduce memory networks
for factoid QA. Memory network framework is extended
in \cite{weston2015towards,kumar2015ask} for Facebook bAbI
dataset. Peng et al. \shortcite{PengLLW15}'s Neural
Reasoner infers over multiple supporting facts to
generate an entity answer for a given question and it
is also tested on bAbI. All of these works
deal with some short texts with simple-grammar, aiming
to \emph{generate} an answer which is restricted to be one
word denoting a location, a person etc.

Some works also tried over other kinds of QA tasks. For
example, Iyyer et al., \shortcite{iyyer2014neural} present
QANTA, a recursive neural network, to infer an entity based
on its description text. This task is basically a matching
between description and entity, no explicit question exist.
Another difference with us lies in that all the sentences in
the entity description actually contain partial information
about the entity, hence a description is supposed to have
only one representation. However in our task, the modeling
of \passage{} should be dynamically changed according to the
question analysis. Hermann et
al., \shortcite{hermann2015teaching} incorporate attention
mechanism into LSTM for a QA task over news text. Still,
their work does not handle some complex question types like
``Why...'', they merely aim to find the entity from
the \passage{} to fill the slot in the query so that the
completed query is true based on
the \passage{}. Nevertheless, it inspires us to treat
our task as a textual entailment problem by first
reformatting question-answer pairs into statements.

Some other deep learning systems are developed for answer
selection
task \cite{yu2014deep,yang2015wikiqa,severyn2015learning,shen2015question,wang2010modeling}. Differently,
this kind of question answering task does not
involve \passage{} comprehension. They only try to match the
question and answer candidate without any background
information. Instead, we treat machine comprehension in this work 
as a question-answer matching problem under
background guidance.

Overall, for open-domain MCTest machine comprehension task, this work is the first to resort to deep neural networks.

\section{Model}\label{sec:model}
We investigate this task by three approaches, illustrated in
Figure \ref{fig:routes}. (i) We can  compute two
different document (D) representations in a common space,
one based on question (Q) attention, one based on answer (A)
attention, and compare them. 
This architecture we name HABCNN-QAP. (ii) 
We compute a  representation of D based on Q
attention (as before), but now we compare it directly with a
representation of A. We name this architecture HABCNN-QP.
(iii) We treat this QA task as textual entailment (TE), first
reformatting Q-A pair into a statement (S), then
matching S and D directly. 
This architecture we name HABCNN-TE.
All three approaches are implemented in the common framework
HABCNN.

\subsection{HABCNN}
\seclabel{habcnn}
Recall that we use the abbreviations A (answer), Q
(question), S (statement), D (document).
HABCNN performs representation learning for triple (Q,
A, D) in HABCNN-QP and HABCNN-QAP, for tuple
(S, D) in HABCNN-TE. For convenience, we use
``query'' to refer to Q, A, or S uniformly. HABCNN, depicted
in Figure \ref{fig:abhcnn}, has the following phases.

\textbf{Input Layer.} 
The input is (query,D). Query is two individual
sentences (for Q, A) or one single sentence (for S),
D is a sequence of sentences. Words are
initialized by $d$-dimensional pre-trained word embeddings. As
a result, each sentence is represented as a feature map with
dimensionality of $d \times s$ ($s$ is sentence length). In
Figure \ref{fig:abhcnn}, each sentence in the input layer is
depicted by a rectangle with multiple columns.

\textbf{Sentence-CNN.} Sentence-CNN is used for sentence representation learning from word level. Given a sentence  of length $s$ with a word sequence: $v_1,v_2,\ldots,v_s$,
let vector $\mathbf{c}_i\in\mathbb{R}^{wd}$ be the
concatenated embeddings of $w$ words
$v_{i-w+1},\ldots,v_{i}$ where $w$ is the filter width, $d$
is the dimensionality of word representations and $0< i
<s+w$.  Embeddings for words $v_i$, $i<1$ and $i>s$, are
zero padding.  We then generate the representation
$\mathbf{p}_i\in\mathbb{R}^{d_1}$ for the \textit{phrase}
$v_{i-w+1},\ldots,v_{i}$ using the convolution weights
$\mathbf{W}\in\mathbb{R}^{d_1\times wd}$:
\begin{equation}
\mathbf{p}_i=\mathrm{tanh}(\mathbf{W}\cdot\mathbf{c}_i+\mathbf{b})
\end{equation}
where bias $\mathbf{b}\in\mathbb{R}^{d_1}$. $d_1$ is called ``kernel size'' in CNN.

Note that the sentence-CNNs for query and all document sentences share the same weights, so that the resulting sentence representations are comparable.

\textbf{Sentence-Level Representation.}
The sentence-CNN generates a new feature map (omitted in
Figure \ref{fig:abhcnn}) for each input sentence, one
column in the feature map denotes a phrase representation
(i.e., $\mathbf{p}_i$ in Equation (1)).

For the query and each sentence of D, we do \emph{element-wise
1-max-pooling} (``max-pooling'' for
short) \cite{collobert2008unified} over phrase
representations to form their representations at this level.

We now treat D as a set of ``vital'' sentences
and ``noise'' sentences. We propose \emph{attention-pooling}
to learn the sentence-level representation of D as follows: first
identify vital sentences by computing attention for each
D's sentence  as the cosine similarity between the
its representation 
and the query representation, then select the $k$
highest-attention sentences to do max-pooling over
them. Taking Figure \ref{fig:abhcnn} as an example, based on
the output of
sentence-CNN layer, $k=2$ important sentences with blue color are
combined by max-pooling as the sentence-level 
representation $\mathbf{v}_s$ of D; the other -- white-color --
sentence representations are neglected as they have low
attentions. (If $k=all$, attention-pooling returns to the common
max-pooling in \cite{collobert2008unified}.) When the
query is (Q,A), this step will be repeated, once for Q, once
for A, to compute
representations of D at the sentence level 
that are biased with respect to Q and A,
respectively.

\textbf{Snippet-CNN.} 
As the example in Figure \ref{fig:example} shows, to answer
the first question ``why did Grandpa answer the door?'', it
does not suffice to compare this question
only to the  sentence ``Grandpa answered the door with a
smile and welcomed Jimmy inside''; instead, the \snippet{}
``Finally, Jimmy arrived at Grandpa's house and
knocked. Grandpa answered the door with a smile and welcomed
Jimmy inside'' should be used to compare. To this end, it is
necessary to stack another CNN layer, \emph{snippet-CNN}, to learn representations
of \snippet{}s, i.e., units of one or more sentences. Thus, the basic units
input to snippet-CNN (resp.\ sentence-CNN) are sentences (resp.\
words)
and the output is representations of snippets (resp.\ sentences).

Concretely, snippet-CNN puts all sentence representations 
in column sequence as a feature map and
conducts another convolution operation over it. With
filter width $w$, this step generates
representation of \snippet{} with $w$ consecutive
sentences. Similarly, we use the same CNN to learn
higher-abstraction query representations (just treating query as a document with only one sentence, so that the higher-abstraction query representation is in the same space with corresponding snippet representations).

\textbf{Snippet-Level Representation.}
For the output of snippet-CNN, each representation is
more abstract and denotes bigger granularity. 
We apply the same attention-pooling process to snippet level representations:
attention values are computed as cosine similarities between query and snippets
and the snippets with the $k$ largest attentions are retained.
Max-pooling over the $k$ selected snippet representations
then creates the snippet-level  representation
$\mathbf{v}_t$ of D. Two
selected \snippet{}s are shown  as red in Figure \ref{fig:abhcnn}.

\textbf{Overall Representation.}
Based on convolution layers at two different granularity, we
have derived query-biased representations of D at
sentence level (i.e., $\mathbf{v}_s$) as well as \snippet{}
level (i.e., $\mathbf{v}_t$).  In order to create a flexible
choice for open Q/A, we develop a highway
network \cite{srivastava2015training} to combine the two
levels of representations as an overall 
representation $\mathbf{v}_o$ of D:

\begin{equation}
\mathbf{v}_o=(1-\mathbf{h})\odot\mathbf{v}_s+\mathbf{h}\odot\mathbf{v}_t
\end{equation}
where highway network weights $\mathbf{h}$ are learned by 
\begin{equation}
\mathbf{h}=\sigma(\mathbf{W}_h\mathbf{v}_s+\mathbf{b})
\end{equation}
where $\mathbf{W}_h\in\mathbb{R}^{d_1\times d_1}$. With the
same highway network, we can generate the overall query
representation, $\mathbf{r}_i$ in Figure \ref{fig:abhcnn},
by combining the two representations of the query at sentence
and snippet levels.

\subsection{HABCNN-QP \& HABCNN-QAP}
HABCNN-QP/QAP computes the representation of D as a
projection of D, either based on attention from Q or based on
attention from A. We hope that these two projections of the
document are close for a correct A and less close for an
incorrect A.
As we said in related
work, machine comprehension can be viewed as an answer
selection task \emph{using the document D as critical
background information}. Here, HABCNN-QP/QAP
do not compare Q and A directly, but they use Q and A to
filter the document differently, extracting what is critical
for the Q/A match by attention-pooling.
Then they match the two document representations in the
new space.

For ease of exposition, we have used the symbol
$\mathbf{v}_o$ so far, but 
in HABCNN-QP/QAP we compute two different document representations:
$\mathbf{v}_{oq}$, for which
attention is computed with respect to Q; and
$\mathbf{v}_{oa}$ for which attention is computed with
respect to
A. 
$\mathbf{r}_i$ also has two versions, one for Q: $\mathbf{r}_{iq}$, one for A: $\mathbf{r}_{ia}$. 

HABCNN-QP and HABCNN-QAP make different use of
$\mathbf{v}_{oq}$.  HABCNN-QP compares $\mathbf{v}_{oq}$
with answer representation $\mathbf{r}_{ia}$.  HABCNN-QAP
compares $\mathbf{v}_{oq}$ with
$\mathbf{v}_{oa}$. HABCNN-QAP projects D twice, once based
on attention from Q, once based on attention from A
and compares the two
projected representations, shown in Figure \ref{fig:routes}
(top).  HABCNN-QP only utilizes the Q-based projection of D 
and then compares the projected document with
the answer representation, shown in Figure \ref{fig:routes}
(middle).

\subsection{HABCNN-TE}
HABCNN-TE treats machine comprehension as \emph{textual
entailment}. We use the statements that are provided as part of
MCTest. Each statement corresponds to 
a question-answer pair; e.g.,
the Q/A pair ``Why
did Grandpa answer the door?'' /  
``Because he saw the insects'' 
(Figure \ref{fig:example})
is
reformatted into the statement ``Grandpa answered the door
because he saw the insects''.
The question answering task is then cast as: ``does the document
entail the statement?''

For HABCNN-TE, shown in Figure \ref{fig:routes} (bottom),
the input for Figure \ref{fig:abhcnn} is the pair (S,D).
HABCNN-TE tries to match the  S's representation $\mathbf{r}_i$ with the
D's representation  $\mathbf{v}_o$.

\section{Experiments}
\subsection{Dataset}
MCTest\footnote{http://requery.microsoft.com/mct} has two
subsets. MCTest-160 is a set of 160 items, each consisting
of a document, four questions followed by one correct anwer and three incorrect answers
(split into
70 train, 30 dev and 60 test) and  MCTest-500 a
set of 500  items (split into 300 train, 50 dev and 150
test).

% The stories are fictional so the answers can be found only
% in the story itself.The stories and questions are
% carefully limited, thereby minimizing the world knowledge
% required for this task. Yet, the task is challenging for
% most modern NLP system. Each story in MCTest has four
% multiple choice questions, each with four answer
% choices. Each question has only one correct
% answer. Furthermore, questions are also annotated with
% ``single'' and ``multiple'' labels. The questions
% annotated ``single'' only require one sentence in the
% story to answer them. For ``multiple'' questions it should
% not be possible to find the answer to the question in any
% individual sentence of the D. In a sense, the
% ``multiple'' questions are harder than the ``single''
% questions as they typically require complex lexical
% analysis, some inference and some form of limited
% reasoning. Cucerzan-converted questions can also be
% downloaded from the MCTest website.

\subsection{Training Setup and Tricks}
Our training objective is to minimize the following ranking loss function:
\begin{equation}
L(d,a^+, a^-)=\mathrm{max}(0, \alpha+S(d,a^-)-S(d,a^+))
\end{equation}
where $S(\cdot,\cdot)$ is a matching score between two representation vectors. Cosine similarity is used throughout. $\alpha$ is a constant.

For this common ranking loss, we also have two styles to utilize the data in view of each positive answer is accompanied with three negative answers. One is treating ($d$, $a^+$, $a^-_1$, $a^-_2$, $a^-_3$) as a training example, then our loss function can have three ``max()'' terms, each for a positive-negative pair; the other one is treating ($d$, $a^+$, $a^-_i$) as an individual training example. In practice, we find the second way works better. We conjecture that the second way has more training examples, and positive answers are repeatedly used to balance the amounts of positive and negative answers.  

\textbf{Multitask learning}: Question typing is commonly
used and proved to be very helpful in QA tasks \cite{sachan2015learning}.
 Inspired, we stack a logistic regression layer over question representation $\mathbf{r}_{iq}$, with the purpose that this subtask can favor the parameter tuning of the whole system, and finally the question is better recognized and is able to find the answer more accurately.

To be specific, we classify questions into  12
classes: ``how'', ``how much'', ``how many'', ``what'',
``who'', ``where'', ``which'', ``when'', ``whose'', ``why'',
``will'' and ``\emph{other}''. The question label is
created by querying for the label keyword in the question.
If more than one keyword appears in a
question, we adopt the one appearing earlier and the more
specific one (e.g., ``how much'', not ``how''). In case
there is no match, the class
``\emph{other}'' is assigned.

We train with AdaGrad \cite{duchi2011adaptive} and
use 50-dimensional GloVe \cite{pennington2014glove} to
initialize word
representations,\footnote{http://nlp.stanford.edu/projects/glove/}
kept fixed during training. 
Table \ref{tab:hyper} gives hyperparameter
values, tuned on dev.
\begin{table}[t]
\setlength{\abovecaptionskip}{0pt} 
\setlength{\belowcaptionskip}{-5pt}
\begin{center}
\setlength{\tabcolsep}{2mm}
\begin{tabular}{ccccccc}
 $k$& lr & $d_1$ & bs & $w$ & $L_2$ & $\alpha$\\\hline
 [1,3]&0.05 & [90, 90] & 1 & [2,2] & 0.0065 & 0.2\\
%  I(two) & 0.085 & [50, 50] & 1 & [3,3] & 0.0003\\\hline
%  II(one) & 0.085 & [50,  --] & 1 & [3,--] & 0.0001\\
%  II(two) & 0.085 & [50, 50] & 1 & [3,3] & 0.0001\\\hline
%  III(one) & 0.05 & [50,  --] & 1 & [3,--] & 0.0003\\
%  III(two) & 0.055 & [50, 50] & 1 & [3,3] & 0.0005\\
\end{tabular}
\end{center}
\caption{Hyperparameters. $k$: top-$k$ in attention-pooling for both CNN layers; lr: learning rate; $d_1$: kernel size in CNN layers; bs: mini-batch size; $w$: filter width; $L_2$: $L_2$ normalization; $\alpha$: constant in loss function.}\label{tab:hyper} 
\end{table}
\begin{table*}[htbp]
\setlength{\abovecaptionskip}{0pt} 
\setlength{\belowcaptionskip}{-10pt} 
\begin{center}
\begin{tabular}{ll|rrr|rrr|rrr|rrr}
&&\multicolumn{6}{c|}{MCTest-150}  & \multicolumn{6}{c}{MCTest-500} \\ 
\multicolumn{2}{c|}{method} &  \multicolumn{3}{c|}{acc}  & \multicolumn{3}{c|}{NDCG$_4$} &\multicolumn{3}{c|}{acc}  & \multicolumn{3}{c}{NDCG$_4$} \\ 
& & one&mul&all & one&mul&all&one&mul&all & one&mul&all\\\hline
\multirow{4}{*}{\rotatebox{90}{Baselines}}&Addition & 39.3 & 32.4 & 35.7 & 60.4 & 50.3 & 54.6 & 35.7 & 30.2 &32.9 & 56.6 & 55.2 & 55.8\\
&Addition-proj & 42.1 & 38.7  & 40.3 & 65.3 & 61.3 & 63.2 & 39.4 & 36.7& 38.0 & 63.3& 60.1& 61.7\\
&AR & 48.1  & 44.7 & 46.3 & 70.5 & 68.9 & 69.6 & 44.4 &39.5 &41.9 & 66.7 & 64.2&65.4\\ 
& NR & 48.4 & 46.8 & 47.6 & 70.7 & 68.2 & 69.7 &45.7 & 45.6 &45.6 & 71.9 & 69.5 &70.6\\\hline
\multicolumn{2}{c|}{Variant-I}&50.4 & 47.7 & 49.0& 73.9& 71.2& 72.5 & 45.4& 42.1&43.7&68.8&64.5&66.6\\
\multicolumn{2}{c|}{Variant-II}& 53.6 & 51.0 & 52.2 & 74.6 & 70.1 & 72.3&47.1 &46.0&46.5 &70.2&63.8&66.9\\\hline
\multicolumn{2}{c|}{HABCNN-QP}&57.9 & 53.7 & 55.7& 80.4& 80.0& 80.2 & 53.7& 46.7&50.1&75.4&72.7&74.0\\
\multicolumn{2}{c|}{HABCNN-QAP}& 59.0 & 57.9 & 58.4 & 81.5 & 79.9 & 80.6&54.0 &47.2&50.6 &75.7&72.6&74.1\\
\multicolumn{2}{c|}{HABCNN-TE}& \textbf{63.3} & \textbf{62.9}& \textbf{63.1}&\textbf{86.6} & \textbf{85.9}& \textbf{86.2}&\textbf{54.2}&\textbf{51.7}&\textbf{52.9}&\textbf{76.1}&\textbf{74.4}&\textbf{75.2}\\
\end{tabular}
\end{center}
\caption{Experimental results for one-sentence (one),
multiple-sentence (mul) and all cases.}\label{tab:overallresult} 
\end{table*}

\begin{figure*}[t]
\setlength{\abovecaptionskip}{0pt} 
\setlength{\belowcaptionskip}{-5pt} 
\centering
\includegraphics[width=14cm]{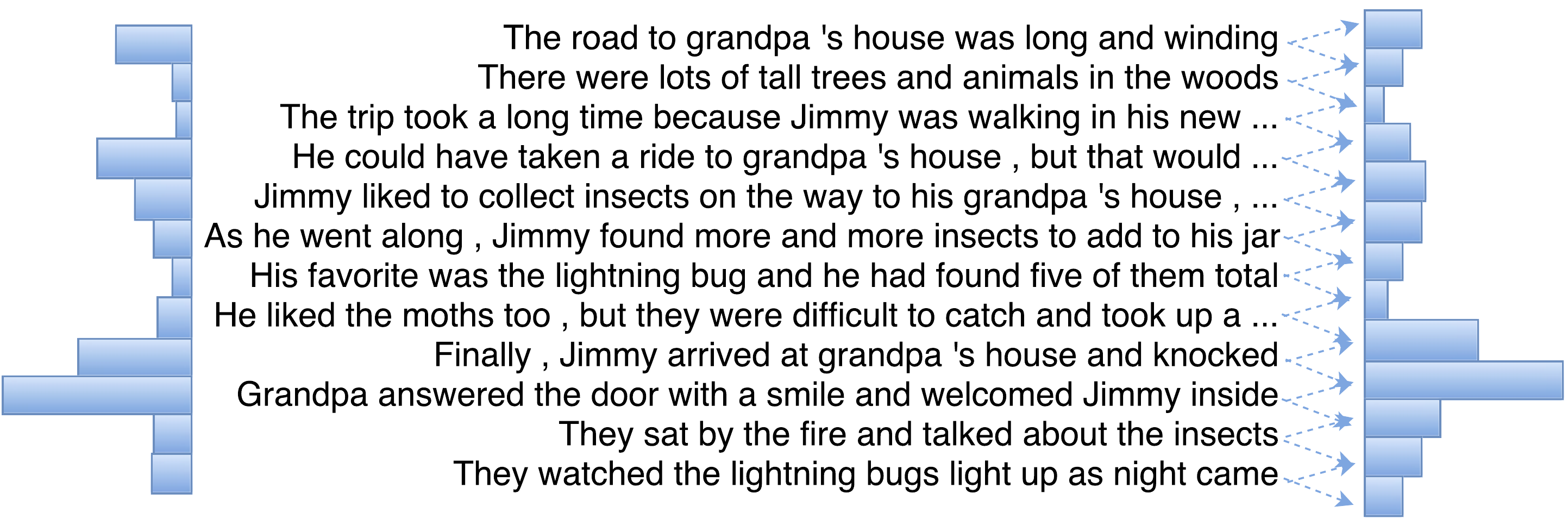}
\caption{Attention visualization for statement ``Grandpa answered the door because Jimmy knocked'' in the example Figure \ref{fig:example}. Too long sentences are truncated with ``$\ldots$''. Left is attention weights for each single sentence after the sentence-CNN, right is attention weights for each \snippet{} (two consecutive sentences as filter width $w=2$ in Table \ref{tab:hyper}) after the snippet-CNN. } \label{fig:visualization}
\end{figure*}

We consider two evaluation metrics: \emph{accuracy} (proportion of questions correctly answered) and \emph{NDCG$_4$} \cite{jarvelin2002cumulated}. Unlike accuracy which evaluates if the question is correctly answered or not, NDCG$_4$, being a measure of ranking quality, evaluates the position of the correct answer in our predicted ranking. 

\subsection{Baseline Systems}
This work focuses on the comparison with systems about distributed representation learning and deep learning:

% \textbf{RTE}: For this baseline, MCTest is re-casted as a RTE task by converting each question-answer pair into a statement (using \cite{cucerzan2005factoid}) and then selecting the answer whose statement has the highest likelihood of being entailed by the story. 

\textbf{Addition.} Directly compare question and answers without considering the D. Sentence representations are computed by element-wise addition over word representations. 

\textbf{Addition-proj.} First compute sentence
representations for Q, A and all D sentences as the same way as \emph{Addition}, then match the two sentences in D which have highest similarity with Q and A respectively.

\textbf{NR.} 
The Neural Reasoner \cite{PengLLW15} has an encoding layer,
multiple reasoning layers and a final answer layer. The
input for the encoding layer is a question and the sentences
of the document (called facts); each
sentence is encoded by a GRU into a vector. In each reasoning
layer, NR lets the question representation interact with
each fact representation as reasoning process. Finally,
all temporary reasoning clues are pooled as answer
representation.

\textbf{AR.} 
The Attentive Reader \cite{hermann2015teaching} is
implemented by modeling the whole D as a word
sequence -- without specific sentence /
\snippet{} representations -- using an LSTM.
Attention mechanism is
implemented at word representation level.

Overall, baselines Addition and Addition-proj do not involve complex
composition and inference. NR and AR represent the
top-performing deep neural networks in QA tasks.

% \textbf{QANTA}: \cite{iyyer2014neural} TQANTA uses a recursive neural network for question answering.

\subsection{HABCNN Variants}
In addition to the main architectures described above, we
also explore 
two variants of ABCHNN, inspired by \cite{PengLLW15} and  \cite{hermann2015teaching}, respectively.

\textbf{Variant-I}: 
As RNNs are widely recognized as a competitor of CNNs in
sentence modeling, similar with \cite{PengLLW15}, we replace
the sentence-CNN in Figure \ref{fig:abhcnn}  by a GRU while keeping other parts unchanged.

\textbf{Variant-II}: 
How to
model attention
at the granularity of \emph{words}
was shown in
\cite{hermann2015teaching}; see their paper for details.
We develop their attention idea and model attention at the
granularity of \emph{sentence and snippet}.
Our attention gives different
weights to  sentences/snippets (not words), then 
computes the document representation as a weighted average
of all sentence/snippet representations.

\subsection{Results}
Table \ref{tab:overallresult} lists the performance of
baselines, HABCNN-TE variants, HABCNN systems in the first,
second and last block, respectively (we only report variants
for top-performing HABCNN-TE).  Consistently, our HABCNN
systems outperform all baselines, especially surpass the two
competitive deep learning based systems AR and NR. 
The
margin between our best-performing ABHCNN-TE and NR is
15.6/16.5 (accuracy/NDCG) on MCTest-150 and 
7.3/4.6 on 
MCTest-500. This
demonstrates the promise of our architecture in this task.

As said before, both AR and NR systems aim to \emph{generate} answers in  entity form.  Their designs might not suit this machine comprehension task, in which the answers are openly-formed based on summarizing or abstracting the clues. To be more specific, AR models D always at word level, attentions are also paid to corresponding word representations, which is applicable for entity-style answers, but is less suitable for comprehension at sentence level or even \snippet{} level.  NR contrarily models D  in sentence level always, neglecting the discovering of key phrases which however compose most of answers. In addition, the attention of AR system and the question-fact interaction  in NR system both bring large numbers of parameters, this potentially constrains their power in a dataset of limited size.

For Variant-I and Variant-II (second block of
Table \ref{tab:overallresult}), we can see that both
modifications do harm to the original HABCNN-TE
performance. The first variant, i.e, replacing the
sentence-CNN in Figure \ref{fig:abhcnn} as GRU module is not
helpful for this task. We suspect that this lies in the
fundamental function of CNN and GRU. The CNN models a
sentence without caring about the global word order
information, and max-pooling is supposed to extract the
features of key phrases in the sentence no matter where the
phrases are located. This property should be useful for
answer detection, as answers are usually formed by
discovering some key phrases, not all words in a sentence
should be considered. However, a GRU models a sentence by
reading the words sequentially, the importance of phrases is
less determined by the question requirement. The second
variant, using a more complicated attention scheme to model
biased D representations than simple cosine
similarity based attention used in our model, is less
effective to detect truly informative sentences
or \snippet{}. We doubt such kind of attention scheme when
used in sentence sequences of large size. In training, the
attention weights after softmax normalization have actually
small difference across sentences, this means the system can
not distinguish key sentences from noise sentences effectively. Our
cosine similarity based attention-pooling, though pretty
simple, is able to filter noise sentences more effectively,
as we only pick top-$k$ pivotal sentences to form D
representation finally. This trick makes the system simple
while effective.

\subsection{Case Study and Error Analysis}
In Figure \ref{fig:visualization}, we visualize the attention distribution at sentence level as well as \snippet{} level for the statement `` Grandpa answered the door because Jimmy knocked'' whose corresponding question requires multiple sentences to answer. From its left part, we can see that ``Grandpa answered the door with a smile and welcomed Jimmy inside'' has the highest attention weight. This meets the intuition that this sentence has semantic overlap with the statement. And yet this sentence does not contain the answer. Look further the right part, in which the CNN layer over sentence-level representations is supposed to extract high-level features of \snippet{}s. In this level, the highest attention weight is cast to the best \snippet{} ``Finally, Jimmy arrived...knocked. Grandpa answered the door...''. And the neighboring \snippet{}s also get relatively higher attentions than other regions.  Recall that our system chooses the one sentence with top attention at left part and choose top-3 \snippet{}s at right part (referring to $k$ value in Table \ref{tab:hyper}) to form D representations at different granularity, then uses a highway network to combine both representations as an overall D representation. This visualization hints that our architecture provides a good way for a question to compromise key information from different granularity.

We also do some preliminary error analysis. One big obstacle for our systems is the ``how many'' questions. For example, for question ``how many rooms did I say I checked?'' and the answer candidates are four digits ``5,4,3,2'' which never appear in the D, but require the counting of some locations. However, these digital answers can not be modeled well by distributed representations so far. In addition, digital answers also appear for ``what'' questions, like ``what time did...''. Another big limitation lies in ``why'' questions. This question type requires complex inference and long-distance dependencies. We observed that all deep lerning systems, including the two baselines, suffered somewhat from it.

\section{Conclusion}
This work takes the lead in presenting a CNN based neural network system for open-domain machine comprehension task. Our systems tried to solve this task in a \passage{} projection way as well as a textual entailment way. The latter one demonstrates slightly better performance. Overall, our architecture, modeling dynamic document representation by attention scheme from sentence level to \snippet{} level, shows promising results in this task. In the future, more fine-grained representation learning approaches are expected to model complex answer types and question types.

\bibliographystyle{named}
\bibliography{ijcai16}

\end{document}